\documentclass[journal,twoside,web]{IEEEtran}
\usepackage{amsmath,amssymb,amsfonts}
\usepackage{algorithmicx,algorithm}
\usepackage{algpseudocode}
\usepackage{array}
\usepackage{textcomp}
\usepackage{stfloats}
\usepackage{url}
\usepackage{verbatim}
\usepackage{graphicx}
\usepackage{cite}

\usepackage{mathrsfs}
\usepackage{cancel}
\usepackage{booktabs}
\usepackage{multirow}
\usepackage{pdfpages}
\usepackage{gensymb} 

\DeclareMathOperator{\Cov}{\mathrm{Cov}}

\begin{document}

\title{A Deep Probabilistic Flow-Based Framework for Unsupervised Cross-Domain Soft Sensing}

\author{Junn Yong Loo$^{1,*}$, Hwa Hui Tew$^{1,*}$, Fang Yu Leong$^{1}$, Ze Yang Ding$^{2}$,\\ Vishnu Monn Baskaran$^{1}$, Chee-Ming Ting$^{1}$ and Chee Pin Tan$^{2,\dagger}$
\thanks{$^{1}$The authors are with the School of Information Technology, Monash University Malaysia, Jalan Lagoon Selatan, Bandar Sunway, 47500 Selangor, Malaysia.
$^{2}$The authors are with the School of Engineering, Monash University Malaysia, Jalan Lagoon Selatan, Bandar Sunway, 47500 Selangor, Malaysia.
$^{*}$These authors contributed equally to this work.
$^{\dagger}$Corresponding authors (Email: tan.chee.pin@monash.edu).}
%
}


\maketitle


\begin{abstract}
Industrial soft sensing is crucial for accurate process monitoring through reliable inference of dominant sensor variables. However, developing effective data-driven soft sensor models presents challenges, such as achieving domain adaptability, addressing incomplete sensor labels, and learning stochastic data variability. To overcome these challenges, we propose a Deep Variational Potential Flow (DVPF) framework for cross-domain soft sensor modeling, taking into account the lack of sensor labels in the target domain. Our framework introduces sequential variational Bayes with recurrent neural network (RNN) parameterization to address the maximum likelihood estimation problem that characterizes cross-domain soft sensing. Central to the framework is a potential flow that performs unsupervised Bayesian inference on the RNN-extracted features to obtain an exact representation of the intractable posterior distribution. Together, these DVPF components learn domain-adaptable features that effectively capture complex cross-domain process dynamics and data variability. We validate the proposed DVPF on a real industrial multiphase flow process across varying operating modes. The results show that the DVPF demonstrates superior performance in cross-domain soft sensing compared to existing deep feature-based domain adaptation methods.
\end{abstract}


\begin{figure*}[t]
\centering
\includegraphics[width=1.0\textwidth]{BlockDiagram.pdf}
\caption{Overall diagram of the proposed DVPF. 
The probabilistic RNN is used first to extract prior latent and hidden features $(z_{t},h_{t})$.
The potential flow (\ref{eq:particle_flow}) then performs unsupervised Bayesian inference on these RNN-extracted features $(z_{t},h_{t})$. The dashed outlines of $y^{\mathcal{S}}_{t}$ indicate that sensor labels are only available within the source domain.} 
\label{fig:block}
\end{figure*}

\section{Introduction} \label{sect:Introduction}

\IEEEPARstart{T}{he} rapid evolution in industrial instrumentation over recent years has driven an increasing demand for managing complex industrial processes characterized by nonlinear dynamics \cite{Sui}. Accurate estimation of both the representations of these dynamic processes and their operating modes is often crucial for effective monitoring and control \cite{Ding}. Soft sensors address these needs as a cost-effective and low-maintenance alternative to hardware sensors, which are often prohibitively expensive or unreliable for dense deployment, especially in harsh environments. Moreover, they provide the high-frequency, spatially distributed data essential for advanced process control, which is frequently infeasible with physical instrumentation alone \cite{Liang,Yang2}. Beyond these benefits, soft sensors also enhance system robustness by providing redundant virtual measurements that can cross-validate physical sensors, detect faults, and improve overall reliability and safety \cite{Yongfang}. Since the advent of deep learning, neural networks have been extensively employed as data-driven soft sensing models \cite{Shen}. While these soft sensors prioritize model generalization by learning meaningful features that capture distinct process dynamics \cite{DLSS1}, they often assume that the sensor data originates from a single domain with a common distribution \cite{Yang}. 
In particular, this assumption of a unified domain is impractical in real-world applications, as modern industrial processes exhibit distinct dynamics and data variability across varying operating modes, thereby representing different process domains\cite{Shi},\cite{Jin}\cite{He2}. Consequently, soft sensor models trained on data acquired under specific operating modes (source domain) struggle to generalize effectively to target applications with unseen operating modes (target domain). Furthermore, acquiring sensor labels across various operating modes is often labor-intensive. In many cases, collecting representative sensor labels may be prohibitively expensive or even unattainable \cite{Rozantsev},\cite{Tang}, resulting in missing labels in the target domain.

To overcome mismatches between the source and target domains, and the challenge of incomplete sensor labels, unsupervised domain adaptation has demonstrated immense potential in leveraging unlabeled target data to generalize deep learning models beyond the labeled source data \cite{Kouw}. Numerous deep feature-based unsupervised domain adaptation models have been proposed for cross-domain time series analysis across various important applications \cite{Wilson}. 
These domain adaptation models predominantly rely on techniques, such as domain distribution alignment \cite{DengWanxia,YangShuai,Shuang} and domain adversarial training \cite{Costa,ChaiTASE,ChenZhuyun}.
Despite the success of these unsupervised domain adaptation approaches, several key challenges remain.
In particular, domain adversarial training, as implemented in Domain-Adversarial Neural Networks (DANN) \cite{Ganin} and related approaches, relies on adversarial competition between the data encoder and a domain discriminator to generate domain-invariant feature representation. However, this approach fundamentally assumes consistent label distributions across domains, which often does not hold in soft sensing applications. When target domain labels are scarce and process relationships vary between domains, enforcing strict domain invariance can inadvertently align features with mismatched label distributions. This introduces source-domain bias and ultimately limits cross-domain generalization performance. Consequently, in the absence of target labels, reliance on domain-invariant features and the assumption of no label shift can lead to feature representations that are heavily biased toward the source labels, thereby limiting model adaptation to the target domain \cite{Kouw}. Conversely, distribution alignment methods employ metrics such as maximum mean discrepancy, correlation alignment loss, and central moment discrepancy to align the distributions of domain-specific models that are independently trained on the source and target domains. However, applying these alignment strategies to domain-exclusive models falls short in capturing the complex cross-domain process dynamics that occur during domain transitions \cite{Purushotham}.

Recently, deep probabilistic approaches that integrate variational Bayes have demonstrated superior capabilities in stochastic data modeling and predictive uncertainty quantification \cite{Ruimin}, \cite{Wu}. The inherent capacity of the variational Bayes framework for generative modeling provides probabilistic solutions that are particularly well-suited for industrial applications characterized by process uncertainties and incomplete data \cite{ChenJiahong,QianDong}. 
Additionally, several studies \cite{Purushotham,TuYouzhi,ChaiTNNLS} have combined sequential variational Bayes with domain adversarial training to improve domain adaptation performance in time-series applications. However, these approaches constrain the variational posterior to a simple parametric family, such as Gaussian or mean-field distributions, which limits representational capacity and produces highly localized posterior estimates \cite{Kingma_3}, thereby reducing generalization capability and hindering effective domain adaptation.
Furthermore, while existing soft sensor models can extract rich temporal features \cite{YuanXiaofeng}, these features are not subject to domain adaptation, thereby restricting the models' generalization across domains \cite{ChaiTNNLS}. Moreover, these soft sensing approaches do not address the problem of missing labels in the target domain.

In this work, we propose a Deep Variational Potential Flow (DVPF) framework to overcome the critical challenge of cross-domain soft sensing in industrial processes, where the absence of target sensor labels and the presence of nonstationary, multi-mode data render traditional models ineffective. To address this, the DVPF framework integrates sequential variational Bayes with a novel unsupervised particle flow, referred to as potential flow. This integration enables the learning of an exact posterior feature representation that accurately characterizes complex cross-domain dynamics. 
Unlike existing approaches, the DVPF employs a Gaussian prior only as an initial proposal and performs a nonlinear, flow-based Bayesian update to obtain a richer, non-Gaussian posterior representation, thereby improving generalization and cross-domain adaptability.
Additionally, our approach abstains from domain adversarial training to mitigate bias toward the source domain due to missing labels in the target domain.
Our framework is specifically designed for multinode processes exhibiting nonlinear dynamics and domain heterogeneity, making it applicable to a wide range of industrial scenarios with multiple operating modes, such as multiphase flow systems, chemical reactors, and manufacturing lines, where it enables accurate soft sensing in unlabeled target domains without deploying additional physical sensors. To the best of our knowledge, this is the first work to integrate variational Bayes with particle flow for unsupervised domain adaptation in soft sensor modeling.

\section{Methods}

To formulate our DVPF framework, we first reframe unsupervised cross-domain soft sensing as a sequential variational inference problem, resulting in a new variational Bayes loss function. Next, we introduce an RNN-based parameterization of this loss function to obtain a fixed-dimensional feature representation, enabling generative feature sampling. Finally, we propose the potential flow to perform Bayesian inference on RNN-extracted features to acquire an exact posterior representation. Fig. \ref{fig:block} summarizes the overall DVPF framework.

\subsection{Cross-Domain Sequential Variational Bayes} \label{ssect:smooth-VBO}

Consider the sequence of source and target data $\{\mathcal{X}^\mathcal{S}, \mathcal{X}^\mathcal{T}\}$ and label $\{\mathcal{Y}^\mathcal{S}, \mathcal{Y}^\mathcal{T}\}$ spaces. Here, $\mathcal{S}$ and $\mathcal{T}$ denote the source and target domains, respectively, with each domain characterized by a distinct joint distribution, i.e., $p(\mathcal{X}^\mathcal{S}, \mathcal{Y}^\mathcal{S}) \neq p(\mathcal{X}^\mathcal{T}, \mathcal{Y}^\mathcal{T})$.
Unsupervised cross-domain soft sensing aims to predict the missing target labels ${y}_{0:T} = \{{y}_{t} \in \mathcal{Y}^\mathcal{T}\}_{t=0}^{T}$, given the domain-agnostic data ${x}_{0:T} = \{{x}_{t} \in \mathcal{X}^\mathcal{S} \cup \mathcal{X}^\mathcal{T}\}_{t=0}^{T}$
and the partially available source labels ${y}^\mathcal{S}_{0:T} = \{{y}_{t} \in \mathcal{Y}^\mathcal{S}\}_{t=0}^{T}$. Here, $t$ denotes the time step and $T$ denotes the sequence length. This unsupervised time series modeling problem is fundamentally a maximum likelihood estimation problem, addressed by maximizing the joint log-likelihood of the data and source labels, i.e., $\max \log p(x_{0:T},y^{\mathcal{S}}_{0:T})$

Given the challenges in directly maximizing this log-likelihood, we introduce the latent sequence ${z}_{0:T} = \{{z}_{t} \in \mathcal{Z}\}_{t=0}^{T}$ and apply importance-weighted decomposition to rewrite the maximum log-likelihood problem as the following maximization–expectation problem:
\begin{align} \label{eq:DPF_joint}
\begin{split}
\max \log p(x_{0:T},y^{\mathcal{S}}_{0:T}) &= \max \mathbb{E}_{q(z_{0:T})}\!\! \left[\log \frac{p(x_{0:T},y^{\mathcal{S}}_{0:T},z_{0:T})}{q(z_{0:T})}\right] \\
&\quad+ \min \mathcal{D}^\text{KL}\big[ q(z_{0:T})\|p(z_{0:T}|x_{0:T}) \big] \\
\end{split}
\end{align}
where $\mathcal{D}^\text{KL}$ denotes the Kullback–Leibler divergence (KLD), and we assume that the posterior distribution $p(z_{0:T}|x_{0:T}) = p(z_{0:T}|x_{0:T},y^{\mathcal{S}}_{0:T})$ of the latent features depends solely on data. This importance decomposition holds for any choice of variational distribution $q(z_{0:T})$.
By applying ancestral factorization to the first expected log-likelihood term in (\ref{eq:DPF_joint}), we obtain
\begin{align} \label{eq:DPF_conditional}
\begin{split}
&\mathbb{E}_{q(z_{0:T})} \left[\log \frac{p(x_{0:T},y^{\mathcal{S}}_{0:T},z_{0:T})}{q(z_{0:T})}\right] \\
&= \sum_{t=0}^{T} \, \mathbb{E}_{q(z_{0:t})} \left[\log \frac{p(x_{t},y^{\mathcal{S}}_{t},z_{t}|x_{0:t-1},y^{\mathcal{S}}_{0:t-1},z_{0:t-1})}{q(z_{t}|z_{0:t-1})}\right] \\
\end{split}
\end{align}
Subsequently, we consider a factorization of the conditional data likelihood in (\ref{eq:DPF_conditional}) as follows:
\begin{align} \label{eq:ancestral_factorization}
\begin{split}
&p(x_{t},y^{\mathcal{S}}_{t},z_{t}|x_{0:t-1},y^{\mathcal{S}}_{0:t-1},z_{0:t-1}) \\
&= p(x_{t}|z_{0:t}) \; p(y^{\mathcal{S}}_{t}|z_{0:t-1}) \; p(z_{t}|z_{0:t-1})
\end{split}
\end{align}
assuming that the latent features are representative of the data.
Finally, by substituting (\ref{eq:ancestral_factorization}) into (\ref{eq:DPF_conditional}), the joint log-likelihood (\ref{eq:DPF_joint}) can then be expressed as the following sequential variational Bayes loss (sVBO) function:
\begin{align} \label{eq:sVBO_objective}
\begin{split}
&\max \mathcal{L}^\text{sVBO}(p,q) = \sum_{t=0}^{T} \, \max \mathcal{L}^\text{sVBO}(p,q;{t}), \\
&\mathcal{L}^\text{sVBO}(p,q;{t}) = \mathbb{E}_{q(z_{0:t})} \big[ \log p(x_{t}|z_{0:t}) \big] \\
&\qquad\qquad\qquad\;\;+ \mathbb{I}_{\mathcal{S}}(y_{t}) \; \mathbb{E}_{q(z_{0:t-1})} \big[ \log p(y_{t}|z_{0:t-1}) \big] \\
&\qquad\qquad\qquad\;\;- \mathcal{D}^\text{KL}\big[ q(z_{t}|z_{0:t-1})\|p(z_{t}|z_{0:t-1}) \big] \\
&\qquad\qquad\qquad\;\;- \mathcal{D}^\text{KL}\big[ q(z_{0:t})\|p(z_{0:t}|x_{0:t}) \big]
\end{split}
\end{align}
where $\mathbb{I}_{\mathcal{S}}(y_{t})$ denotes the indicator function, i.e., we consider the label expected log-likelihood only when labels ${y}_{t} \in \mathcal{Y}_{t}^\mathcal{S}$ are available within the source domain.
Given that the KLD is non-negative, we express its minimization as the maximization of its negative. In particular, our sVBO incorporates the second expected log-likelihood to account for the semi-supervision provided by the partial (source) labels. Furthermore, our sVBO formulation produces the intractable KLD of the joint distributions, which performs Bayesian smoothing of the latent state sequence. This endows the latent features with a posterior distribution that effectively captures the temporal characteristics of the sensor data across domains. 

Additionally, instead of minimizing the full-sequence KLD $\mathcal{D}^{\mathrm{KL}}\big[q(z_{0:T})\|p(z_{0:T}|x_{0:T})\big]$ only at the terminal timestep $T$, which requires traversing the entire sequence and backpropagating through long temporal horizons, we minimize the partial-sequence KLD $\mathcal{D}^{\mathrm{KL}}\big[q(z_{0:t})\|p(z_{0:t}|x_{0:t})\big]$ iteratively across time steps.
By applying the chain rule to the full-sequence KLD, we obtain the following decomposition:
\begin{align} \label{eq:chain_rule_decomposition}
\begin{split}
&\; \mathcal{D}^{\mathrm{KL}}[q(z_{0:T})\|p(z_{0:T}| x_{0:T})]
= \mathcal{D}^{\mathrm{KL}}[q(z_{0:t})\|p(z_{0:t}| x_{0:t})] \\ 
&+ \mathbb{E}_{q(z{0:t})}\big[\mathcal{D}^{\mathrm{KL}}\big[q(z_{t+1:T}|z_{0:t}) \| p(z_{t+1:T}|z_{0:t},x_{0:T})\big]\big] \\
&+
\mathbb{E}_{q(z_{0:t})}\bigg[\log\frac{p(z_{0:t}|x_{0:t})}{p(z_{0:t}|x_{0:T})}\bigg]
\end{split}
\end{align}
Notably, the second term inside the expectation is a non-negative KLD, and when a minimum of the partial-sequence KLD is attained at $q(z_{0:t}) = p(z_{0:t}| x_{0:t})$, the last term likewise reduces to a non-negative KLD $\mathcal{D}^{\mathrm{KL}}\big[p(z_{0:t}| x_{0:t})\|p(z_{0:t}| x_{0:T})\big] \geq 0$.
This KLD measures the divergence between the filtered posterior $p(z_{0:t}| x_{0:T})$ and a smoothed posterior $p(z_{0:t}| x_{0:T})$, which decreases as the memory-encoding hidden state $h_t$, summarizing the historical information $z_{0:t-1}$, is updated with additional observations over time.
Hence, the partial-sequence KLD serves as a principled surrogate for the full-sequence KLD, yielding a time-local training objective that is naturally compatible with the recurrent parameterization and enables the model to receive gradient feedback throughout the sequence rather than only at the terminal timestep.

Consequently, we reformulate unsupervised cross-domain soft sensing as a variational inference problem described by the sVBO (\ref{eq:sVBO_objective}). 
In particular, the expected log-likelihoods of the sVBO aim to derive the generative conditional distributions $p(x_{t}|z_{0:t})$ and $p(y^{\mathcal{S}}_{t}|z_{0:t-1})$ to enable sequence modeling of the data and the partial source labels. The KLD losses, on the other hand, aim to match the variational approximations $q(z_{t}|z_{0:t-1})$ and $q(z_{0:t})$, respectively to the conditional $p(z_{t}|z_{0:t-1})$ and the intractable joint posterior $p(z_{0:t}|x_{0:t})$ latent distributions. Our reformulation shows that the pair of conditional and joint variational distributions satisfying this KLD matching defines a domain-adaptable latent space $\{\mathcal{Z}_{t}\}_{t=0}^{T}$, facilitating unsupervised cross-domain soft sensing without target labels.

\subsection{Recurrent Neural Network Parameterization} \label{ssect:modified_VRNN}

To enable direct sampling from the set of generative conditional distributions in sVBO (\ref{eq:sVBO_objective}), we first parameterize them as Gaussian distributions, given by:
\begin{subequations} \label{eq:Gaussian_generative_models}
\begin{align}
\label{eq:Gaussian_measurement_likelihood}
&p(x_{t}|z_{0:t}) := p(x_{t}|z_t,h_t) = \mathcal{N} (\mu^{\text{dec}}_{t} , \Sigma^{\text{dec}}_{t}), \\
\label{eq:Gaussian_state_prior}
&p(z_{t}|z_{0:t-1}) := p(z_{t}|h_t) = \mathcal{N} (\mu^{\text{prior}}_{t} , \Sigma^{\text{prior}}_{t})
\end{align}
\end{subequations}
where $\Sigma^{\text{dec}}_{t} = \text{diag}(\sigma_{t}^{\text{dec}^2})$ and $\Sigma^{\text{prior}}_{t} = \text{diag}(\sigma_{t}^{\text{prior}^2})$ are both isotropic covariances, and $\text{diag}$ denotes the diagonal function. 
Here, we coincide the conditional label distribution $p(y^{\mathcal{S}}_{t}|z_{0:t-1})$ with the conditional latent distributions (\ref{eq:Gaussian_state_prior}), allowing these latent distributions to predict the source labels when available, thereby enforcing label supervision of the latent features within the source domain.
Additionally, we parameterize the means and standard deviations of the generative distributions (\ref{eq:Gaussian_generative_models}) via a probabilistic recurrent neural network (RNN), as follows:
\begin{subequations} \label{eq:modified_VRNN}
\begin{align}
\label{eq:modified_VRNN_dec}
&(\mu^{\text{dec}}_{t} , \sigma^{\text{dec}}_{t}) = \varphi_{\theta}^{\text{dec}} (z_{t},h_{t}), \\
\label{eq:modified_VRNN_prior}
&(\mu^{\text{prior}}_{t} , \sigma^{\text{prior}}_{t}) = \varphi_{\theta}^{\text{prior}} (h_{t}), \\
\label{eq:modified_VRNN_rnn}
&h_{t} = \varphi_{\theta}^{\text{rnn}} (z_{t-1}, h_{t-1})
\end{align}
\end{subequations}
where both the models $\varphi_{\theta}^{\text{prior}}$ and $\varphi_{\theta}^{\text{dec}}$ are implemented as feed-forward neural networks. 
In particular, $\varphi_{\theta}^{\text{rnn}}$ represents the RNN, and $h_{t} \in \mathcal{H}$ denotes its memory-encoding hidden features, which are responsible for capturing the intrinsic temporal information from the historical data.

The adoption of an RNN parameterization is motivated by its recurrent structure, which yields fixed-dimensional features comprising the latent and hidden features $(z_{t}, h_{t}) \in \mathcal{Z} \times \mathcal{H}$. These RNN-extracted features provide an efficient augmentation of the latent state sequence, whose dimensionality increases over time steps.
Applying this RNN parameterization to the sVBO (\ref{eq:sVBO_objective}) yields
\begin{align} \label{eq:sVBO_objective_rnn}
\begin{split}
&\mathcal{L}^\text{sVBO}(\theta;{t}) = \mathbb{E}_{q(z_{t},h_{t})}\big[ \log p(x_{t}|z_{t},h_{t}) \big] \\
&\qquad\qquad\qquad\!\!+ \mathbb{I}_{\mathcal{S}}(y_{t}) \; \mathbb{E}_{q(h_{t})}\big[ \log p(y_{t}|h_{t}) \big] \\
&\qquad\qquad\qquad\!\!- \mathcal{D}^\text{KL}\big[ q(z_{t}|h_{t})\|p(z_{t}|h_{t}) \big] \\
&\qquad\qquad\qquad\!\!- \mathcal{D}^\text{KL}\big[ q(z_{t},h_{t}) \| p^+(z_{t},h_{t}) \big]
\end{split}
\end{align}
where we denote $p^+(z_{t},h_{t}) := p^+(z_{t},h_{t}|x_{0:t})$ as the posterior distribution for brevity.
As such, the RNN parameterization enables ancestral sampling from the transition probability
$p(z_{t},h_{t}|z_{t-1},h_{t-1}) = p(z_{t}|h_{t}) \, p(h_{t}|z_{t-1},h_{t-1})$
which corresponds to the models in (\ref{eq:modified_VRNN_prior}) and (\ref{eq:modified_VRNN_rnn}).
Although the RNN model represents a deterministic mapping, the resulting hidden state samples remain stochastic due to the variability of the preceding-time inputs.
Consequently, the generated samples are associated with the following predictive prior distribution:
\begin{align} \label{eq:prior_distribution}
\begin{split}
&\; p^{-}(z_{t},h_{t}) := p(z_{t},h_{t}|x_{0:t-1}) \\
&= \int \, p(z_{t},h_{t}|z_{t-1},h_{t-1}) \, p^{+}(z_{t-1},h_{t-1}) \; d (z_{t-1},h_{t-1})
\end{split}
\end{align}
provided the input pairs $(z_{t-1},h_{t-1})$ are drawn from the posterior distribution.
For our soft sensing task, we define the conditional prior as $q(z_{t}|h_{t}) = p(z_{t}|h_{t})$ following (\ref{eq:Gaussian_state_prior}), in which case $q(z_{t},h_{t}) = p^{-}(z_{t},h_{t})$ and the first KLD term in the sVBO formulation vanishes.
Prior to the Bayesian update, the latent features have not observed the measurement, since $x_t$ is not incorporated as input to the RNN models.
To obtain an exact posterior representation, i.e., to perform the Bayes update 
$q(z_{t},h_{t}) \rightarrow p^{+}(z_{t},h_{t})$, we subsequently perform Bayesian inference on the latent features via a potential flow, detailed in Section \ref{ssect:VO_DPF_novel}.
We note that the Gaussian model in (\ref{eq:Gaussian_state_prior}) is used solely to parameterize the initial conditional prior $p(z_{t} | h_{t})$, which is standard in variational Bayes to enable tractable sampling and likelihood evaluation. 

\begin{figure}[t]
\centering
\includegraphics[width=1.0\columnwidth]{FlowDiagram.pdf}
\caption{Graphical illustrations of the overall DVPF framework and its operations. The dashed outlines of $y^{\mathcal{S}}_{t}$ indicate that sensor labels are only available within the source domain.}
\label{fig:flow}
\end{figure}

\begin{algorithm}[b]
\caption{DVPF Training} 
\label{algo:DVPF_training}
\begin{algorithmic}
\small
\Require Data $x_{t}$, source labels $y^{\mathcal{S}}_{t}$, sequence length $T$
\State Sample $\bar{z}_{0} \sim \mathcal{N}( \mathbf{0}, \mathrm{I})$ and $\bar{h}_{0} \sim \mathcal{N}( \mathbf{0}, \mathrm{I})$
\For{$t \in \{ 1,\dots,T \}$}
\State Compute $h_{t} \gets \varphi_{\theta}^{\text{rnn}} (\bar{z}_{t-1}, \bar{h}_{t-1})$
\State Compute $(\mu^{\text{prior}}_{t} , \sigma^{\text{prior}}_{t}) \gets \varphi_{\theta}^{prior} (h_{t})$
\State Sample $z_{t} \gets \mu^{\text{prior}}_{t} + \epsilon_{t} \, \sigma^{\text{prior}}_{t}$
\State Compute $\nabla \phi_{\vartheta} (z_{t}, h_{t}; x_{t})$ via backpropagation
\State Update $({z}^+_{t}, {h}^+_{t}) \gets (z_{t}, h_{t}) + \nabla \phi_{\vartheta} (z_{t}, h_{t}; x_{t})$
\State Compute $(\mu^{\text{dec}}_{t} , \sigma^{\text{dec}}_{t}) \gets \varphi_{\theta}^{dec} ({z}^+_{t}, {h}^+_{t})$
\EndFor
\State Compute $\nabla_{\theta,\vartheta} \, \mathcal{L}^\mathrm{sVBO} (\theta,\vartheta)$ based on (\ref{eq:sVBO_final}) via backpropagation
\State Update neural network parameters $\theta, \vartheta$
\end{algorithmic}
\end{algorithm}

\subsection{Deep Variational Potential Flow} \label{ssect:VO_DPF_novel}

To address the KLD loss within the RNN-parameterized sVBO (\ref{eq:sVBO_objective_rnn}), we present a particle flow framework to construct an exact flow approximation of the intractable posterior $p^+(z_{t},h_{t})$. Specifically, we subject samples (particles) of the RNN-extracted features to a series of differentiable transformations (diffeomorphisms). These transformations are designed such that the transformed samples correspond to a variational distribution $q({z}_{t}, {h}_{t})$ that closely approximates the intractable posterior. This inherently performs exact Bayesian inference on the RNN-extracted features to facilitate the learning of a domain-adaptable posterior feature representation.

Inspired by the control-oriented approaches to Bayesian inference \cite{Laugesen,Yang_3,Olmez}, we propose {\em potential flow}, in which particle transformations are guided by the irrotational velocity field generated by a scalar potential (energy).
Our proposed potential flow is governed by an ordinary differential equation (ODE), as follows:
\begin{align} \label{eq:particle_flow}
\frac{d(z_{t}({\tau}),h_{t}({\tau}))}{d\tau} = \nabla \phi(z_{t}({\tau}),h_{t}({\tau}))
\end{align}
defined over a pseudo-time interval $\tau \in [0, 1]$. In particular, $\phi: \mathcal{Z} \times \mathcal{H} \rightarrow \mathbb{R}$ represents the scalar velocity potential, and
$\nabla$ denotes the Del operator (gradient).
Here, $({z}_{t}({\tau}),{h}_{t}({\tau}))$ denotes the continuous trajectory (flow) of the feature samples within the specified ODE interval. 
Consequently, this potential flow gives rise to a time-dependent sample variational distribution $q({z}_{t}({\tau}),{h}_{t}({\tau}))$, whose evolution follows the partial differential equation (PDE):
\begin{align} \label{eq:Kolmogorov_forward_equation}
\frac{\partial q({\tau})}{\partial \tau} = - \, \nabla \cdot \big( q({z}_{t}({\tau}),{h}_{t}({\tau})) \; \nabla \phi({z}_{t}({\tau}),{h}_{t}({\tau})) \big)
\end{align}
This PDE describes how the probability density $q$ evolves over the pseudo-time $\tau \in [0,1]$ under the divergence of the potential-induced velocity field. The evolution governed by this nonlinear PDE is equivalent to a diffeomorphic transport of the initial proposal distribution $q(z_t(0), h_t(0))$ at $\tau=0$ toward a new distribution $q(z_t(1), h_t(1))$ at $\tau=1$. The resulting variational distribution can be interpreted as the pushforward of the initial proposal under this learned nonlinear diffeomorphic map $\mathcal{T}_\phi : z_t(0), h_t(0) \mapsto z_t(1), h_t(1)$. Unless the flow field $\nabla \phi$ is globally affine, the pushforward of a Gaussian under such a nonlinear diffeomorphism is no longer Gaussian. This is the mechanism by which the DVPF avoids the high-localization limitation of standard Gaussian, mean-field variational approximations.


Subsequently, we formulate a variational potential loss function to perform Bayesian inference on the RNN-extracted features. 
It has been shown in \cite[Proposition 2]{Yang_3} and \cite[Proposition 3]{TMLR} that the evolution of a density homotopy $\rho(z_{t}({\tau}),h_{t}({\tau}))$, which bridges the prior distribution to the posterior distribution, is governed by the following PDE:
\begin{align}
\begin{split} \label{eq:log_homotopy}
\frac{\partial \rho({\tau})}{\partial \tau} = \rho(z_{t}({\tau}),h_{t}({\tau})) \, \big( \Gamma(z_{t}({\tau}),h_{t}({\tau});x_t) - \hat{\Gamma} \big)
\end{split}
\end{align}
where $\Gamma$ denotes the normalized innovation squared:
\begin{align} \label{eq:normalized_innovation_squared}
\begin{split}
&\Gamma(\cdot;x_t) = \frac{1}{2} \, ( x_{t} - \mu^{\text{dec}}_{t}(\cdot) )^{T} {\Sigma_{t}^{\text{dec}}(\cdot)}^{-1} ( x_{t} - \mu^{\text{dec}}_{t}(\cdot) )
\end{split}
\end{align}
with $(\mu^{\text{dec}}_{t},\Sigma_{t}^{\text{dec}})$ obtained from (\ref{eq:Gaussian_measurement_likelihood}), and $\hat{\Gamma} = \mathbb{E}[\Gamma]$.
To facilitate exact Bayesian inference via the posterior update $p^{-}(z_{t},h_{t}) \rightarrow p^{+}(z_{t},h_{t})$, we match the density evolution of the flow-driven variational (\ref{eq:Kolmogorov_forward_equation}) to that of the homotopy (\ref{eq:log_homotopy}), yielding the following generalized Poisson equation:
\begin{align}
\begin{split} \label{eq:Proposition_1_PDE}
&\;\nabla \cdot \big( q({z}_{t}({\tau}),{h}_{t}({\tau})) \, \nabla \phi ( {z}_{t}({\tau}),{h}_{t}({\tau}) ) \big) \\
&= q({z}_{t}({\tau}),{h}_{t}({\tau})) \, \big( \Gamma({z}_{t}({\tau}),{h}_{t}({\tau});x_t) - \hat{\Gamma} \big)
\end{split}
\end{align}
As demonstrated in \cite[Theorem 4.2]{Laugesen}, solving this generalized Poisson equation yields a minimum intractable KLD:
\begin{align}
\begin{split} \label{eq:min_KLD_problem}
\mathcal{D}^\text{KL}\big[ q({z}_{t}({\tau}),{h}_{t}({\tau})) \,\big\|\, \rho(z_{t}({\tau}),h_{t}({\tau})) \big]
\end{split}
\end{align} 
between the flow-driven variational and the density homotopy. Notably, at $\tau=1$, we obtain $q({z}_{t}({1}),{h}_{t}({1}))$ which closely approximates $\rho(z_{t}({1}),h_{t}({1})) \equiv p^{+}({z}_{t},{h}_{t})$, thereby addressing the intractable KLD loss in our RNN-parameterized sVBO (\ref{eq:sVBO_objective_rnn}).

Nevertheless, explicitly solving the generalized Poisson equation (\ref{eq:Proposition_1_PDE}) presents challenges in high-dimensional settings, particularly when neural networks are involved. Numerical solutions, such as the Galerkin and diffusion map methods, often do not scale effectively with dimensionality \cite{Olmez}. Taking this into account, we propose a variational (Deep Ritz) formulation of the generalized Poisson equation. Assuming that the velocity potential $\phi \in \mathcal{H}^1_0(\mathcal{Z} \times \mathcal{H};q)$ belongs to the Sobolev space, it can be demonstrated, via the Calculus of Variations \cite[Chapter 8]{PDE}, that minimizing the following energy functional:
\begin{align} \label{eq:DF_based_functional}
\begin{split}
&\mathcal{L}^\text{VAPO}(\phi;{t}) = 
\mathbb{E}_{q({z}_{t}({\tau}),{h}_{t}({\tau}))} \big[ \big\| \nabla \phi({z}_{t}({\tau}),{h}_{t}({\tau})) \big\|^{2} \big] \\
&+ \Cov_{q({z}_{t}({\tau}),{h}_{t}({\tau}))} \big[ \phi({z}_{t}({\tau}),{h}_{t}({\tau})) , \Gamma({z}_{t}({\tau}),{h}_{t}({\tau});x_t) \big] 
\end{split}
\end{align}
with respect to $\phi$, solves the weak form of the generalized Poisson equation (\ref{eq:Proposition_1_PDE}). Here, $\|\cdot\|$ denotes the Euclidean norm, and $\Cov$ denotes the covariance.
This variational formulation allows us to cast the intractable KLD (\ref{eq:min_KLD_problem}) into a tractable variational potential (VAPO) loss function (\ref{eq:DF_based_functional}). 

Notably, the covariance within this loss enforces the velocity potential to be inversely proportional to the normalized innovation squared (\ref{eq:normalized_innovation_squared}), which quantifies the accuracy of data reconstruction from the conditional data likelihood (\ref{eq:Gaussian_measurement_likelihood}). Since gradients always point in the direction of steepest potential ascent, a minimum covariance ensures that the potential-generated velocity field $\nabla \phi$ consistently guides the RNN-extracted feature samples toward high-likelihood regions characterized by the intractable posterior, thereby inherently performing Bayesian inference. 
Additionally, the VAPO loss function (\ref{eq:DF_based_functional}) does not depend on either the source or target labels. Therefore, our proposed approach represents an unsupervised Bayesian learning paradigm, effectively endowing the RNN features with a domain-adaptable posterior feature representation to achieve cross-domain soft sensing.

In this work, we parameterize the velocity potential $\phi_{\vartheta}$ as a feed-forward neural network with parameters $\vartheta$, ensuring its membership in Sobolev space \cite{Czarnecki}.
Given that the measurement $x_t$ explicitly contributes to the posterior update through $\Gamma$, it is natural to include the measurement as input to the potential function to provide the necessary measurement information. Hence, we incorporate the $x_t$ into the potential function, yielding $\phi_{\vartheta}(z_{t},h_{t}; x_{t})$, to enable a more accurate, data-informed maximum a posteriori update consistent with Bayesian inference.
Furthermore, assuming that the acquired time-series data adequately capture the nonlinear process dynamics with minimal sampling error, we can perform potential flow transformation (\ref{eq:particle_flow}) in forward Euler steps, i.e., the RNN features are updated via $({z}_{t}(\tau + \Delta_{\tau}),{h}_{t}(\tau + \Delta_{\tau})) = ({z}_{t}(\tau),{h}_{t}(\tau)) + \Delta_{\tau} \nabla \phi_{\vartheta}({z}_{t}(\tau),{h}_{t}(\tau); x_{t})$
denotes the flow-transformed feature samples, where $\Delta_{\tau}$ is the Euler step size.
By substituting the intractable KLD loss within our RNN-parameterized sVBO (\ref{eq:sVBO_objective_rnn}) with the VAPO loss function (\ref{eq:DF_based_functional}) and set $\Delta_{\tau} = 1$ (single Euler step), we obtain the final sVBO of our DVPF framework, as follows:
\begin{align} \label{eq:sVBO_final}
\begin{split}
&\mathcal{L}^\text{sVBO}(\theta,\vartheta;{t}) = \mathbb{E}_{q({z}^+_{t},{h}^+_{t})}\big[ \log p(x_{t}|{z}^+_{t},{h}^+_{t}) \big] \\
&\qquad\qquad\quad\!+ \mathbb{I}_{\mathcal{S}}(y_{t}) \; \mathbb{E}_{q({z}^+_{t},{h}^+_{t})}\big[ \log p(y^{\mathcal{S}}_{t}|{h}^+_{t}) \big] \\
&\qquad\qquad\quad\!+ \mathbb{E}_{q({z}_{t},{h}_{t})} \big[ \| \nabla \phi_{\vartheta}({z}_{t},h_{t}; x_{t}) \|^{2} \big] \\
&\qquad\qquad\quad\!+ \Cov_{q(z_{t},h_{t})} \big[ \phi_{\vartheta}(z_{t},h_{t}; x_{t}) , \Gamma(z_{t},h_{t};x_t) \big] 
\end{split}
\end{align}
where $({z}_{t},{h}_{t}) \sim q({z}_{t}(0),{h}_{t}(0))$ and $({z}^+_{t},{h}^+_{t}) \sim q({z}_{t}(1),{h}_{t}(1))$ denote the latent features drawn from the variational distributions, before and after the potential flow (Bayesian inference), respectively.

Overall, our proposed DVPF framework models the generative conditional likelihoods via the probabilistic RNN parameterization, capturing both stochastic uncertainties and temporal patterns underlying the data and partial source labels. In parallel, the potential flow performs unsupervised Bayesian inference on the features extracted by the RNN, yielding a posterior representation that effectively characterizes the domain-agnostic data. Together, these components enable the DVPF to extract domain-adaptable features that accurately characterize the complex dynamics of cross-domain processes. Fig. \ref{fig:flow} provides an overview of the DVPF framework. Training procedures of the DVPF are detailed in Algorithm \ref{algo:DVPF_training}.

\begin{figure}[t]
\centering
\includegraphics[width=1.0\columnwidth]{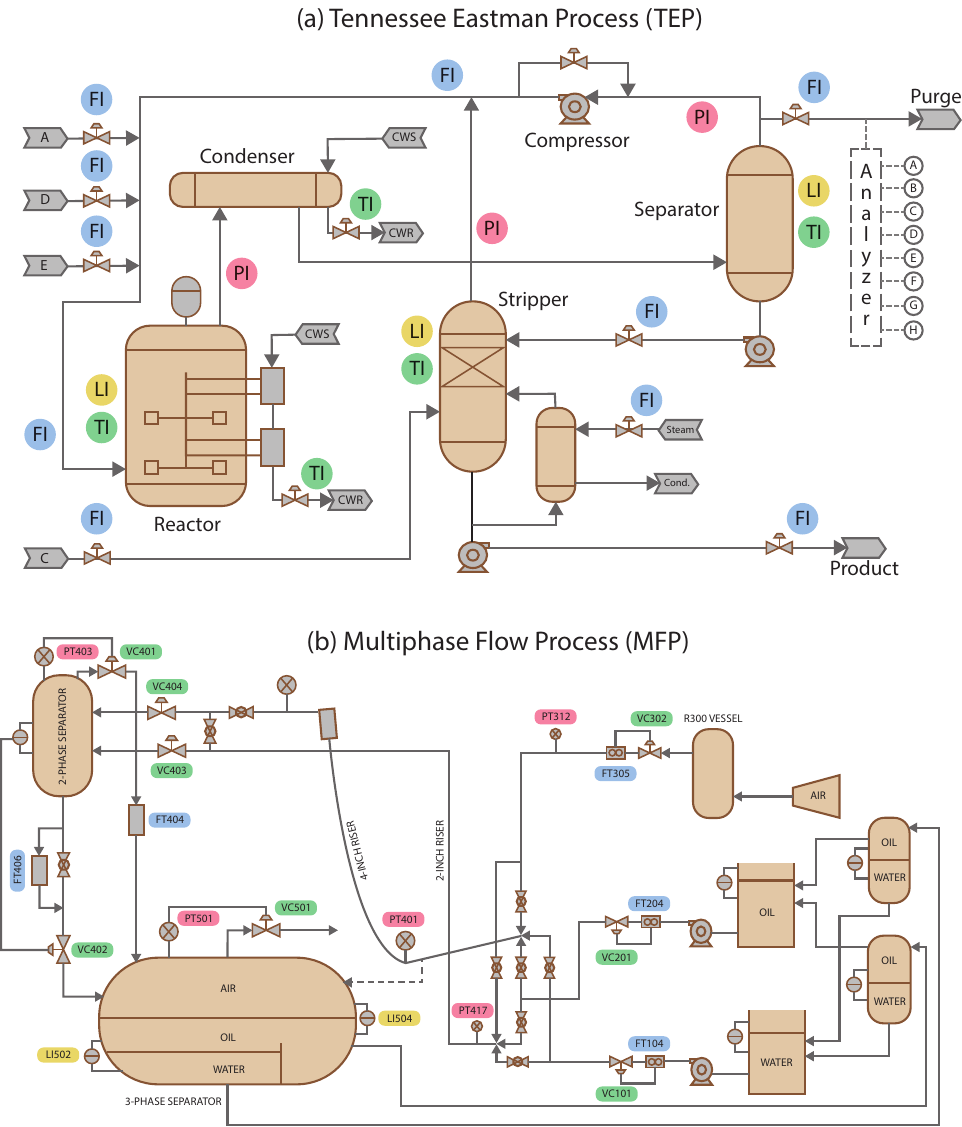}
\caption{Process flow diagrams for the (a) TEP and (b) MFP facilities.}
\label{fig:TEPMFP}
\end{figure}

\section{Case Study I: multiphase flow process (MFP)} \label{sect:Case_StudiesI}
In this section, we present an industrial case study on a multivariate multiphase flow process (MFP), characterized by multiple operating modes and nonlinear dynamics. 
To assess the cross-domain soft sensing performance of the DVPF framework, we validate it against several existing deep feature-based domain adaptation approaches that are based on the various techniques: domain alignment \textbf{DLSN} \cite{DengWanxia}, domain adversarial \textbf{DARNN} \cite{Costa}, and variational Bayes \textbf{MVI} \cite{QianDong}. In addition, we include approaches that combine these techniques: \textbf{VRADA} \cite{Purushotham}, \textbf{InfoVDANN} \cite{TuYouzhi}, and \textbf{DPTR} \cite{ChaiTNNLS}. Furthermore, we extend the ladder-suppression network in \textbf{DLSN} \cite{DengWanxia} to Recurrent Ladder Networks \cite{Isabeau}, enabling time-series modeling.

\subsection{MFP Experiment Setup} \label{sect:MFP}

The Cranfield Multiphase Flow Process (MFP) \cite{Ruiz} is a real industrial process that employs advanced condition monitoring techniques to control the flow of air, water, and oil supplies within its pressurized system. The process flow diagram is provided in Fig. \ref{fig:TEPMFP}.
During data acquisition, the air and water flow rates were continuously varied between set points (operating modes) listed in Table \ref{tab:MFP_setpoints} to achieve a diverse range of process changes, thereby capturing a wide spectrum of distinct dynamics \cite{Wang}.
For our case study, the process variables that necessitate specialized measurement sensors, such as the pressures, flow rates, and densities, are identified as the dominant variables that will be predicted by (output of) the soft sensor. The remaining process variables are considered to be the available sensor measurements, from which the data are input to the soft sensor model.
Table \ref{tab:MFP} lists the input and output process variables of the soft sensor model. Although the target variables considered in our case study can be measured using conventional hardware sensors, soft sensing remains practically relevant in modern process industries \cite{Ma}. Specifically, soft sensors can reduce instrumentation costs in harsh or uninstrumented environments, provide redundancy and validation for physical measurements, and enable high-frequency or spatially distributed monitoring \cite{Yao,He}. These advantages make soft sensing a valuable complement to physical instrumentation, even for commonly measurable variables.


\begin{table}[t]
\centering
\caption{Typical setpoint values for air and water flow rates.}
\label{tab:MFP_setpoints}
\resizebox{\columnwidth}{!}{%
\begin{tabular}{@{}llllll@{}}
\toprule
Input water flow rate & 0.5    & 1      & 2      & 3.5    & 6   \\ \midrule
Input air flow rate & 0.0208 & 0.0278 & 0.0347 & 0.0417 &  \\ 
\bottomrule
\end{tabular}%
}
\end{table}

\begin{table}[t]
\centering
\caption{TEP operation modes of Mass Ratio and Production Rate.}
\label{tab:TEP_setpoints}
\resizebox{\columnwidth}{!}{%
\begin{tabular}{@{}llllllll@{}}
\toprule
Mode & 1 & 2 & 3 & 4 & 5 & 6  \\ 
\midrule
Desired G/H Mass Ratio & 50/50 & 10/90 & 90/10 & 50/50 & 10/90 &90/10  \\ 
\midrule
Desired Production Rate (kg/h) & 14076 & 14077  & 11111 & Max & Max & Max \\ 
\bottomrule
\end{tabular}%
}
\end{table}

\begin{table}[t]
\centering
\caption{The list of MFP variables.}
\label{tab:MFP}
\resizebox{\columnwidth}{!}{%
\begin{tabular}{@{}lclll@{}}
\toprule
Variables & No. & Locations & Descriptions \\ \midrule
\multirow{7}{*}{\begin{tabular}[l]{@{}l@{}}Soft sensor\\input (data)\end{tabular}} & 1 & LI405 & Level of two-phase separator \\
& 2  & FT104        & Input water density \\
& 3  & LI504        & Three-phase separator level \\
& 4  & VC501        & Valve VC501 position \\
& 5  & VC302        & Valve VC302 position \\
& 6  & VC101        & Valve VC101 position \\
& 7  & PO1          & Water pump current\\ \midrule
\multirow{13}{*}{\begin{tabular}[l]{@{}l@{}}Soft sensor\\output (labels)\end{tabular}} & 1 & PT312 & Input air pressure \\
& 2  & PT401        & Bottom riser pressure \\
& 3  & PT408        & Top riser pressure \\
& 4  & PT403        & Two-phase separator pressure \\
& 5  & PT501        & Three-phase separator pressure \\
& 6  & PT408        & Differential pressure (over PT401-PT408) \\
& 7  & PT403        & Differential pressure (over VC404) \\
& 8  & FT305        & Input air flow rate \\
& 9  & FT104        & Input water flow rate \\
& 10  & FT407        & Top riser flow rate \\
& 11 & FT406        & Two-phase separator flow rate \\
& 12 & FT407        & Top riser density \\
& 13 & FT406        & Three-phase separator density \\ \bottomrule
\end{tabular}%
}
\end{table}

\begin{table*}[t]
\centering
\caption{Overall results of cross-domain soft sensing of DVPF and baselines on the TEP dataset. Results of soft sensing are obtained for two bidirectional transfer tasks with independent model training and testing runs.}
\label{tab:MFP_CaseStudies_results}
\resizebox{\textwidth}{!}{%
\setlength\extrarowheight{0.1cm}
\begin{tabular}{@{}c|rrr|rrr|rrr|rrr|rrr|rrr@{}}
\toprule
Variables 
& \multicolumn{18}{c}{Soft Sensor Output No. 1 to 13 (refer Table \ref{tab:MFP} and \cite{Ruiz})} \\ 
\midrule
Tasks
& \multicolumn{9}{c|}{Input Air Flow Rates: Source \{0.0278, 0.0347\} $\rightarrow$ Target \{0.0208, 0.0417\}} 
& \multicolumn{9}{c}{Input Air Flow Rates: Source \{0.0208, 0.0417\} $\rightarrow$ Target \{0.0278, 0.0347\}} \\
\midrule 
Metrics 
& \multicolumn{3}{c|}{RMSE} 
& \multicolumn{3}{c|}{R2} 
& \multicolumn{3}{c|}{MAE}
& \multicolumn{3}{c|}{RMSE} 
& \multicolumn{3}{c|}{R2} 
& \multicolumn{3}{c}{MAE} \\
\midrule 
Domains
& Source & Target & Overall
& Source & Target & Overall
& Source & Target & Overall
& Source & Target & Overall
& Source & Target & Overall
& Source & Target & Overall \\ 
\midrule
DLSN \cite{DengWanxia}
& 0.442 & 0.549 & 0.506
& 0.675 & 0.612 & 0.645
& 0.333 & 0.456 & 0.386
& 0.442 & 0.691 & 0.589
& 0.720 & -0.116 & 0.422
& 0.334 & 0.575 & 0.449 \\
DARNN \cite{Costa}
& 0.390 & 0.626 & 0.537
& 0.685 & 0.532 & 0.595
& 0.284 & 0.471 & 0.381
& 0.455 & 0.795 & 0.650
& 0.690 & -0.385 & 0.388
& 0.324 & 0.586 & 0.450 \\
VRADA \cite{Purushotham}
& 0.376 & 0.570 & 0.497
& 0.697 & 0.592 & 0.632
& 0.278 & 0.436 & 0.360
& 0.435 & 0.493 & 0.477
& 0.708 & 0.340 & 0.639
& 0.319 & 0.383 & 0.350 \\
MVI \cite{ChenJiahong}
& 0.448 & 0.543 & 0.506
& 0.561 & 0.581 & 0.579
& 0.338 & 0.397 & 0.369
& 0.431 & 0.480 & 0.461
& 0.707 & 0.489 & 0.642
& 0.327 & 0.369 & 0.347 \\
InfoVDANN \cite{TuYouzhi}
& 0.474 & 0.557 & 0.525
& 0.486 & 0.547 & 0.531
& 0.372 & 0.406 & 0.389
& 0.400 & 0.451 & 0.432
& 0.755 & 0.573 & 0.703
& 0.293 & 0.354 & 0.318 \\
DPTR \cite{ChaiTNNLS}
& 0.447 & 0.649 & 0.568
& 0.574 & 0.481 & 0.531
& 0.339 & 0.496 & 0.420
& 0.491 & 0.599 & 0.553
& 0.536 & 0.383 & 0.546
& 0.352 & 0.435 & 0.395 \\
\midrule
\textbf{DVPF (ours)}
& \textbf{0.341} & \textbf{0.437} & \textbf{0.401}
& \textbf{0.740} & \textbf{0.725} & \textbf{0.736}
& \textbf{0.243} & \textbf{0.322} & \textbf{0.284}
& \textbf{0.363} & \textbf{0.379} & \textbf{0.377}
& \textbf{0.797} & \textbf{0.698} & \textbf{0.770}
& \textbf{0.262} & \textbf{0.277} & \textbf{0.269} \\
\bottomrule
\end{tabular}
}
\end{table*}

\begin{table}[t]
\centering
\caption{Comparison of Overall 3-fold Cross Validation (CV) Soft Sensing performance. (\text{A|B}: A is train set, B is Test Set). Top: soft sensing metrics; Bottom: Divergence metrics}
\label{tab:Cross_Validation}
\resizebox{\columnwidth}{!}{%
\begin{tabular}{@{}l|c|c|c@{}}
\toprule
\multicolumn{1}{c|}{\multirow{3}{*}{CV Method}} 
& \multicolumn{3}{c}{Soft Sensor Output No. 1 to 13} \\
\cmidrule{2-4}
\multicolumn{1}{c|}{} 
& \multicolumn{1}{c|}{A: T2T3 $|$ B: T1} 
& \multicolumn{1}{c|}{A: T1T3 $|$ B: T2} 
& \multicolumn{1}{c}{A: T1T2 $|$ B: T3} \\ 

\midrule
NRMSE & 0.394  & 0.402  & 0.781 \\ 
R2    & 0.635  & 0.685  & 0.237 \\ 
MAE   & 0.273  & 0.285  & 0.580 \\ 
\midrule
MMD   & 0.098  & 0.167  & 0.274 \\ 
KLD   & 0.032 & 0.139 & 0.246  \\ 
\bottomrule
\end{tabular}%
}
\end{table}

\begin{figure}[ht]
\centering
\includegraphics[width=1.0\columnwidth]{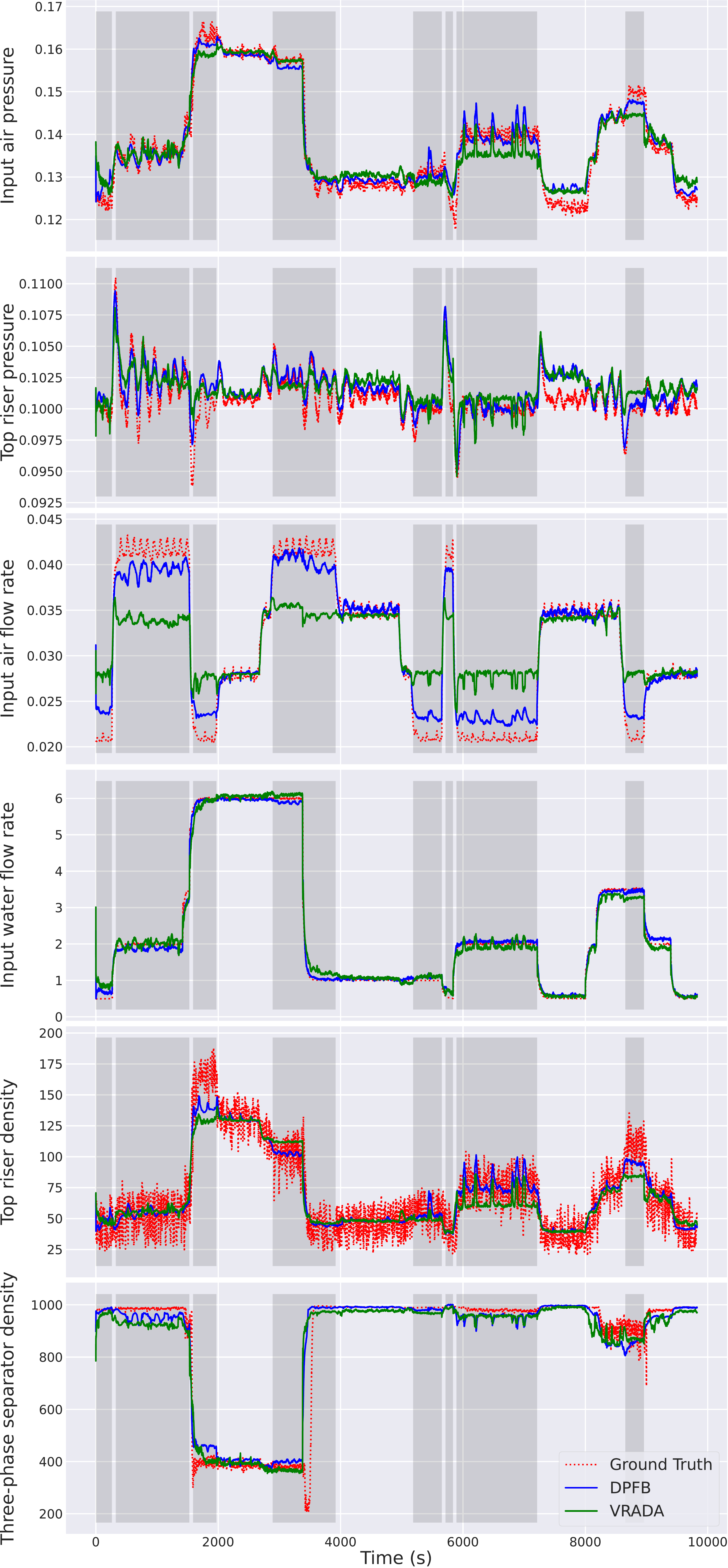}
\caption{Comparisons between ground truths and predictions of the two best models for the first MFP domain adaptation task. The unshaded regions represent the labeled source domain, and the shaded regions represent the unlabeled target domain.}
\label{fig:MFP_results}
\end{figure}

The MFP dataset comprises three time series, each with 13,200, 10,372, and 9,825 data points, respectively. We adopt a 3-fold cross-validation method, using the first two time series for training, while the last is reserved for testing. Notably, the target labels are utilized only as ground truth for result validation.
In this case study, we consider two domain-adaptation tasks defined over the discrete air flow rate setpoints \{0.0208, 0.0278, 0.0347, 0.0417\} m$^3$/s of the MFP benchmark. In the first task, samples collected at the intermediate operating modes \{0.0278, 0.0347\} m$^3$/s are treated as the source domain, accounting for approximately 45\% of the training data, while samples from the remaining setpoints \{0.0208, 0.0417\} m$^3$/s are designated as the target domain, for which labels are assumed to be unavailable. In the second task, the source and target roles are reversed, with \{0.0208,0.0417\} m$^3$/s forming the source domain and \{0.0278,0.0347\} m$^3$/s constituting the target domain. The resulting time-series data (and labels) therefore consist of sequences that alternate between labeled source-domain operating regimes and unlabeled target-domain operating regimes.

For network hyperparameters, we employ models of either a single-layer Gated Recurrent Unit (GRU) $\varphi_{\theta}^{\text{gru}}$, or a Long Short Term Memory (LSTM) $\varphi_{\theta}^{\text{lstm}}$ with a hidden state dimension of $h_{t}^{\text{rnn}} \in \mathbb{R}^{512}$ as the RNN component of our proposed DVPF framework. 
The RNN implementations follow the standard GRU and LSTM architectures as described in [2] and [31], respectively. Specifically, the GRU employs sigmoid activations for the update and reset gates and a tanh activation for the candidate hidden state, whereas the LSTM uses sigmoid activations for the input, forget, and output gates, together with a tanh activation for the cell-state update.
In all baseline methods, the RNN component is implemented as a single-layer LSTM. The models $\varphi_{\theta}^{prior}$ and $\varphi_{\theta}^{dec}$ implemented as feed-forward neural networks with hidden layer dimensions of (256, 128) and (512, 256, 128), respectively, where each entry within the parentheses represents a hidden layer, and its value indicates the dimensionality. The velocity potential function $\phi_{\vartheta}$ is implemented as a bottleneck feed-forward neural network with hidden layers of dimensions (512, 256, 128). These hyperparameters are selected based on validation using the training data.

During training in both case studies, the time-series data and labels are segmented into fixed-length sequences of $L=100$ and grouped into batches. 
Following \cite{ChaiTNNLS}, we optimize all models using the Adam optimizer, but with a larger weight decay of 0.01, and train for a maximum of 300 epochs. Particularly, Adam integrates adaptive learning rates with momentum, combining the advantages of momentum-based optimizers (e.g., stochastic gradient descent with momentum, SGDM) and adaptive optimizers (e.g., RMSprop).
Batch sizes and learning rates are selected from \{8, 16, 32\} and \{5$\times\text{10}^{-4}$, 1$\times\text{10}^{-4}$, 5$\times\text{10}^{-5}$\}, respectively.
In both case studies, we use a batch size of 8 and a learning rate of 5$\times\text{10}^{-4}$, which provides a favorable trade-off between training convergence time, computational resources, and model performance. Notably, we observed that reducing the learning rate leads to substantially longer convergence while yielding no significant improvement in overall performance.
A sequence length of 100 and 8 sample particles are used to compute the empirical approximation of expectations in the sVBO during optimization. At test time, we adopt deterministic inference without reparameterization.

\subsection{MFP Results and Discussions}

To quantify soft sensing performance, we employ the Normalized Root Mean Square Error (NRMSE) and the Coefficient of Determination (R2) to assess the consistency between process variable predictions and ground-truth labels.
As shown in Table \ref{tab:MFP_CaseStudies_results}, our proposed DVPF framework outperforms existing deep feature-based domain adaptation methods, achieving the lowest NRMSE and highest R2 scores both the domain-adaptation directions (\{0.0278, 0.0347\} → \{0.0208, 0.0417\} and \{0.0208, 0.0417\} → \{0.0278, 0.0347\}). 
This comparative analysis provides quantitative evidence supporting our methodological claim that the DVPF consistently outperforms domain adversarial-based methods (DARNN, VRADA, InfoVDANN, DPTR), demonstrating that avoiding adversarial alignment mitigates source-domain bias, particularly under missing target labels.
Notably, the results show consistent relative performance and trends across methods under both domain-adaptation directions, demonstrating a robust cross-domain generalization of the proposed cross-domain framework under different source–target operating configurations.
Most importantly, the performance gains in NRMSE and R2 between the DVPF and the baselines become more pronounced in the target domains, highlighting the DVPF's domain adaptability in achieving accurate soft sensing across alternating source and target domains. Notably, the results shown in Table \ref{tab:Cross_Validation} validate the robustness of the DVPF through a 3-fold cross-validation method, where the framework consistently maintains high prediction accuracy across different time-series segments and unseen process dynamics within the unlabeled target domain. These findings further indicate that the DVPF not only achieve consistent performance across all folds but also outperforms baselines based on normalizing flows or particle filters more effectively with our proposed potential flow.
Fig. \ref{fig:MFP_results} illustrates the soft sensor predictions compared to the ground truth labels. 
Notably, within the target domains, the predictions from our proposed DVPF framework align more closely with the ground truths than the second-best performing VRADA model, particularly evident in the air flow rate predictions (third row of Fig. \ref{fig:MFP_results}). Unlike  the VRADA, the DVPF’s air flow rate predictions are less constrained within the 0.0278 - 0.0347 m$^3$/s source domain range. The ability of the DVPF to generalize its predictions from the labeled source domain to the unlabeled target domains is attributed to the potential flow’s capacity for exact Bayesian inference on the RNN-extracted features. This approach overcomes the limitations of Gaussian and mean-field distributions in variational Bayes, enabling the model to fully leverage the expressiveness of an exact posterior representation to capture nonlinear cross-domain process dynamics and stochastic data variability.

\begin{figure*}[t]
\centering
\includegraphics[width=1.0\textwidth]{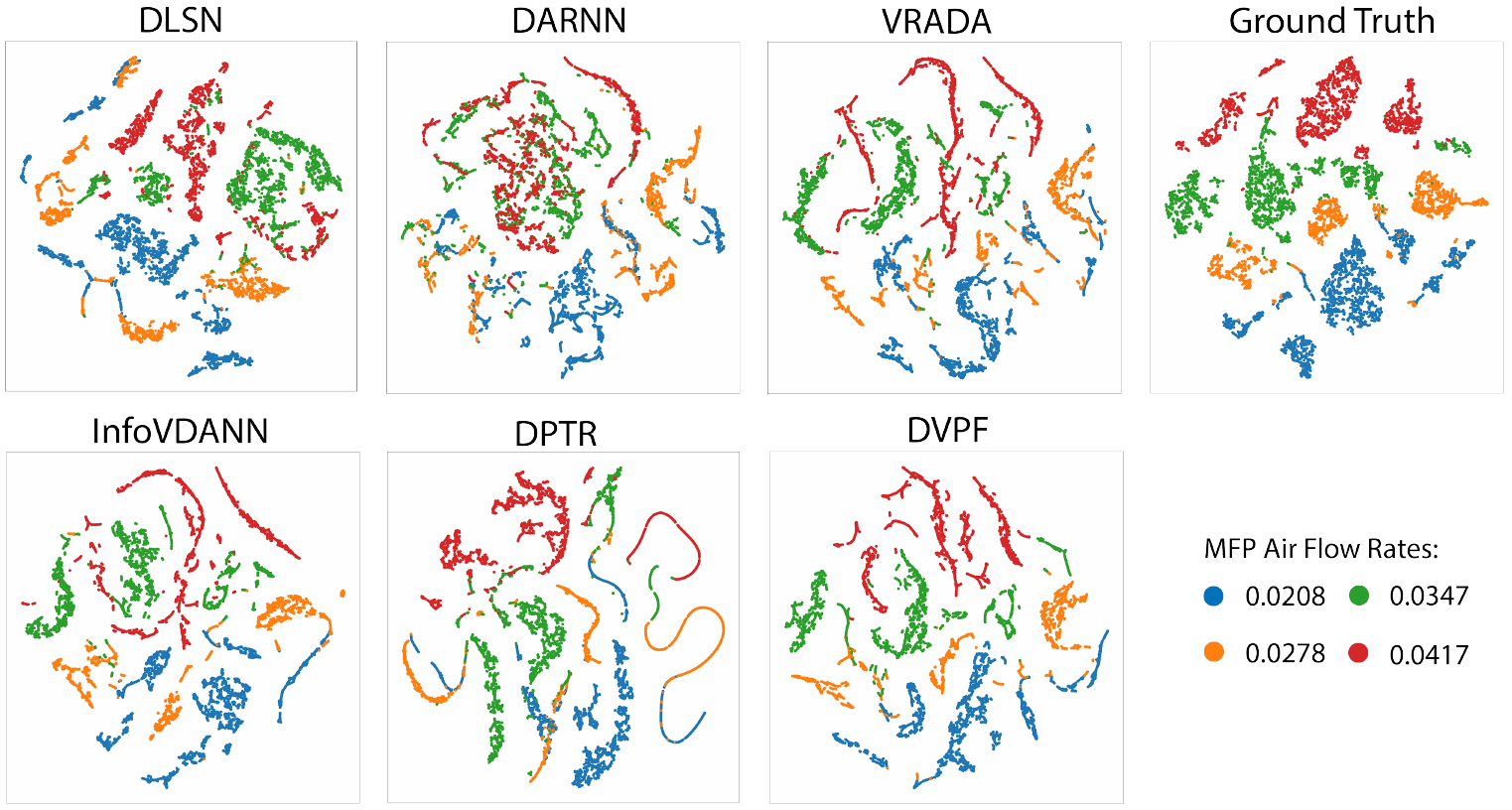}
\caption{Comparison of t-SNE embeddings between the ground-truth sensor labels and the model predictions of the MFP output variables on transfer task: \{0.0278, 0.0347\} → \{0.0208, 0.0417\}.}
\label{fig:tsne_mfp}
\end{figure*}

Fig. \ref{fig:tsne_mfp} shows the t-SNE embeddings of the learned latent features alongside the ground-truth labels. Compared with the baseline methods, the DVPF-learned latent features exhibit more coherent clustering and more closely resemble the ground-truth embedding structure. Moreover, the DVPF demonstrates improved alignment between source and target samples, with reduced overlap across operating domains. These observations indicate that the DVPF learns domain-transferable representations under the complex and highly nonlinear process dynamics of the MFP.
These qualitative observations are consistent with the quantitative results reported in Table \ref{tab:Cross_Validation}, where DVPF achieves lower Maximum Mean Discrepancy (MMD) and Kullback–Leibler Divergence (KLD) values, indicating that the learned posterior features maintain minimal distributional divergence across different operating modes. 
This behaviour can be attributed to the exclusion of domain-invariant feature learning mechanisms within the proposed framework. By avoiding explicit domain-invariance constraints in the absence of target labels, the model mitigates bias toward the labeled source domain while preserving the generalisation capacity of the learned representations, thereby facilitating effective unsupervised cross-domain soft sensing.

\begin{table}[t]
\centering
\caption{Hyperparameter configurations for ablation study on the MFP task: \{0.0278, 0.0347\} → \{0.0208, 0.0417\}.}
\label{tab:MFP_hyperparameter_settings}
\resizebox{\columnwidth}{!}{%
\begin{tabular}{@{}lllllll@{}}
\toprule
Configurations & Model Type & $h_{t}^{\text{rnn}}$ & $\phi_{\vartheta}$ & $f_{\tau}$ &$L$\\ 
\midrule
Setting 1 & GRU   & \{512\} & \{512, 256, 128\} & 1 & 100\\ 
Setting 2 & GRU   & \{512\} & \{512, 256, 128\} & 2 & 100\\ 
Setting 3 & GRU   & \{512\} & \{512, 256, 128\} & 3 & 100\\ 
Setting 4 & LSTM  & \{512\} & \{512, 256, 128\} & 1 & 100\\ 
Setting 5 & LSTM  & \{512\} & \{512, 256, 128\} & 2 & 100\\ 
Setting 6 & LSTM  & \{512\} & \{512, 256, 128\} & 3 & 100\\ 
Setting 7 & GRU   & \{512\} & \{512, 256, 256, 128\} & 1 & 100\\ 
Setting 8 & LSTM  & \{512\} & \{512, 256, 256, 128\} & 1 & 100\\ 
Setting 9 & GRU   & \{256\} & \{256, 128\} & 1 & 100\\ 
Setting 10 & LSTM & \{256\} & \{256, 128\} & 1 & 100\\ 
Setting 11 & GRU & \{512\} & \{512, 256, 128\} & 1 & 50\\ 
Setting 12 & GRU & \{512\} & \{512, 256, 128\} & 1 & 150\\ 
\bottomrule
\end{tabular}%
}
\end{table}

\begin{table}[t]
\centering
\caption{Performance of different hyperparameter settings}
\label{tab:hyperparameter_exp}
\resizebox{\columnwidth}{!}{%
\begin{tabular}{@{}llllllll@{}}
\toprule
Model & Setting 1 & Setting 2 & Setting 3 & Setting 4 & Setting 5 & Setting 6\\ 
\midrule
NRMSE & 0.402 & 0.403 & 0.403 & 0.543 & 0.545 & 0.546\\ 
R2   & 0.686 & 0.685 & 0.684 & 0.194 & 0.193 & 0.193\\ 
MAE  & 0.286 & 0.286 & 0.287 & 0.394 & 0.396 & 0.396\\ 
\midrule
Model & Setting 7 & Setting 8 & Setting 9 & Setting 10 & Setting 11 & Setting 12 \\ 
\midrule
NRMSE  & 0.409 & 0.538 & 0.430 & 0.549 & 0.493 & 0.423 \\ 
R2    & 0.670 & 0.226 & 0.662 & 0.223 & 0.479 & 0.649 \\ 
MAE   & 0.294 & 0.395 & 0.315 & 0.403 & 0.364 & 0.301 \\
\bottomrule
\end{tabular}%
}
\end{table}

\subsection{Hyperparameter Sensitivity Analysis}
Table \ref{tab:MFP_hyperparameter_settings} presents the hyperparameter settings used to evaluate the sensitivity of the proposed DVPF framework. The experiments focus on four key parameters: (i) the RNN model type and hidden-state dimension $h_t$, (ii) the sequence length $L$, (iii) the Euler step frequency $f_\tau = 1/\Delta_{\tau}$, and (iv) the depth and width of the potential-function network $\phi_{\vartheta}$. 
As shown in Table~\ref{tab:hyperparameter_exp}, the GRU-based models consistently outperform the LSTM-based models. The GRU’s reset and update gates provide smoother and more stable latent-state transitions, which align well with the potential-flow updates used in our framework. 
Besides, the model performance is influenced by the choice of sequence length $L$. A larger value of $L$ aggravates vanishing gradients in RNN training, whereas shorter sequences result in insufficient temporal context to capture long-range process dynamics. As shown in Table~\ref{tab:hyperparameter_exp}, a sequence length of $L=100$ (Setting 1) achieves the best trade-off between temporal context and optimization stability, when compared to $L=50$ (Setting 11) and $L=150$ (Setting 12).
Additionally, the experiments with $f_\tau=1$ (Setting 1) to $f_\tau=3$ (Setting 3) indicate that a single-step Euler update is sufficient for accurately approximating posterior evolution in high-dimensional latent spaces. In contrast, the LSTM employs a more complex state-update mechanism with separate input, forget, output, and cell gates. This increased architectural complexity can lead to less stable discretized flow updates under smoothly evolving dynamics, thereby making the model more susceptible to model overfitting.

\begin{figure*}[t]
\centering
\includegraphics[width=1.0\textwidth]{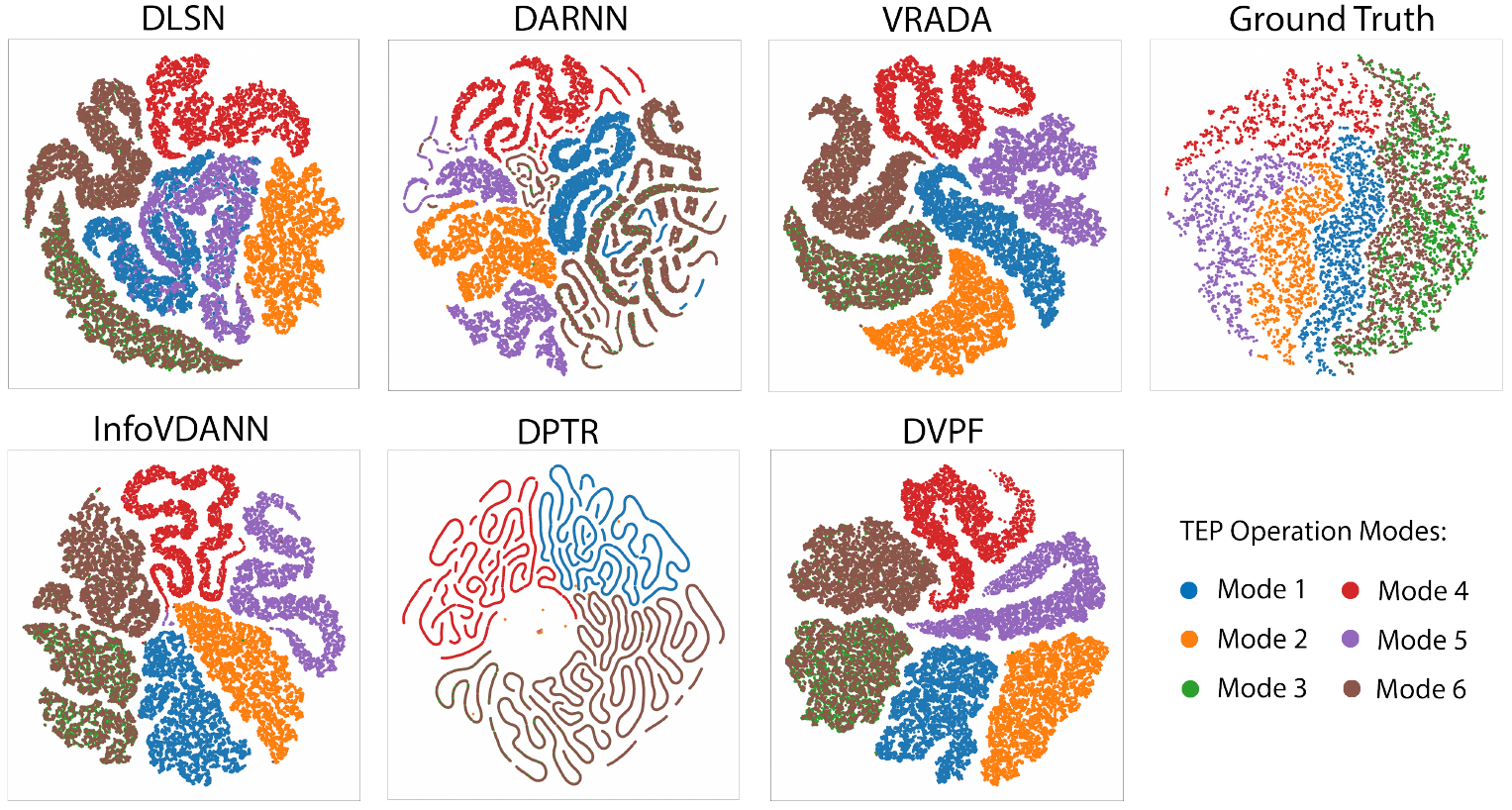}
\caption{Comparison of t-SNE embeddings between the ground-truth sensor labels and the model predictions of the TEP output variables on transfer task: M2, M3, M4, M5, M6 → M1.}
\label{fig:tsne_tep}
\end{figure*}

\section{Case Study II: Tennessee Eastman process} \label{sect:Case_StudiesII}
In this section, we present an additional industrial case study utilizing the publicly available, complex, and large-scale Tennessee Eastman process (TEP). 

\begin{table*}[t]
\centering
\caption{Overall results of cross-domain soft sensing of DVPF and baselines on the TEP dataset. Results of soft sensing are obtained for six leave-one-domain-out transfer tasks with independent model training and testing runs.}
\label{tab:TEP_CaseStudies_results}
\resizebox{\textwidth}{!}{%
\setlength\extrarowheight{0.1cm}
\begin{tabular}{@{}c|ccc|ccc|ccc|ccc|ccc|ccc@{}}
\toprule
Variables 
& \multicolumn{18}{c}{Soft Sensor Output XME 37 to 41 (Product, refer \cite{Montesuma,Reinartz})} \\ 
\midrule
Tasks & \multicolumn{3}{c|}{M2,M3,M4,M5,M6 $\rightarrow$ M1} & \multicolumn{3}{c|}{M1,M3,M4,M5,M6 $\rightarrow$ M2} & \multicolumn{3}{c|}{M1,M2,M4,M5,M6 $\rightarrow$ M3} & \multicolumn{3}{c|}{M1,M2,M3,M5,M6 $\rightarrow$ M4} & \multicolumn{3}{c|}{M1,M2,M3,M4,M6 $\rightarrow$ M5} & \multicolumn{3}{c}{M1,M2,M3,M4,M5 $\rightarrow$ M6} \\ 
\midrule 
Metrics & \multicolumn{1}{c}{RMSE} & \multicolumn{1}{c}{R2} & \multicolumn{1}{c|}{MAE} 
& \multicolumn{1}{c}{RMSE} & \multicolumn{1}{c}{R2} & \multicolumn{1}{c|}{MAE} 
& \multicolumn{1}{c}{RMSE} & \multicolumn{1}{c}{R2} & \multicolumn{1}{c|}{MAE} 
& \multicolumn{1}{c}{RMSE} & \multicolumn{1}{c}{R2} & \multicolumn{1}{c|}{MAE} 
& \multicolumn{1}{c}{RMSE} & \multicolumn{1}{c}{R2} & \multicolumn{1}{c|}{MAE} 
& \multicolumn{1}{c}{RMSE} & \multicolumn{1}{c}{R2} & \multicolumn{1}{c}{MAE} 
\\ 
\midrule
DLSN \cite{DengWanxia} 
& 0.851 & 0.148 & 0.597
& 0.232 & 0.898 & 0.161
& 0.724 & 0.543 & 0.479
& 0.682 & 0.551 & 0.330
& 0.487 & 0.764 & 0.370
& 0.203 & 0.906 & 0.155 \\
DARNN \cite{Costa} 
& 0.498 & 0.688 & 0.296
& 0.512 & 0.737 & 0.297
& 0.540 & 0.728 & 0.307
& 0.627 & 0.609 & 0.413 
& 0.539 & 0.708 & 0.318 
& 0.703 & 0.553 & 0.464 \\
VRADA \cite{Purushotham} 
& 0.478 & 0.718 & 0.267 
& 0.196 & 0.905 & 0.145 
& 0.180 & 0.908 & 0.142 
& 0.721 & 0.409 & 0.476 
& 0.198 & 0.904 & 0.146
& 0.190 & 0.907 &
0.144 \\
InfoVDANN \cite{TuYouzhi} 
& 0.431 & 0.757 & 0.251 
& 0.205 & 0.904 & 0.151 
& 0.182 & 0.908 & 0.139 
& 0.417 & 0.809 & 0.243 
& 0.197 & 0.905 & 0.145
& 0.208 & 0.905 & 0.703 \\
DPTR \cite{ChaiTNNLS} 
& 0.594 & 0.566 & 0.399 
& 0.703 & 0.515 & 0.528 
& 0.211 & 0.905 & 0.161
& 0.816 & 0.354 & 0.415 
& 0.584 & 0.637 & 0.384 
& 0.228 & 0.899 & 0.171 \\
\midrule
\textbf{DVPF} (ours) 
& 0.366 & 0.815 & 0.217 
& 0.208 & 0.903 & 0.153 
& 0.204 & 0.905 & 0.155 
& 0.375 & 0.828 & 0.222 
& 0.174 & 0.908 & 0.136 
& 0.254 & 0.894 & 0.179 \\
\bottomrule
\end{tabular}
}
\end{table*}

\subsection{TEP Experimental Setup}
Fig. \ref{fig:TEPMFP} shows the TEP \cite{Montesuma,Reinartz} flowcharts with five main units: reactor, condenser, stripper, compressor, and separator. The process produces two liquid products (G and H) from four gaseous reactants (A, C, D, and E) and operates under six distinct modes (M1-M6) as illustrated in Table \ref{tab:TEP_setpoints}. 
The process involves a total of 53 variables, including 41 measurement variables (XME) and 12 manipulated variables (XMV). 
We follow the same setup in selecting the 22 continuous measurement variables as auxiliary inputs for predicting the 5 dominant product composition variables, which represent the system's final yield quality \cite{Montesuma,Reinartz,Ding}. 
In this study, only normal operating data are used (consisting of 100 multivariate time series, each containing 600 data points sampled every 3 minutes). 
The proposed DVPF framework is evaluated on the TEP dataset using the same benchmark models as in the MFP study, including \textbf{DLSN} \cite{DengWanxia}, \textbf{DARNN} \cite{Costa}, \textbf{VRADA} \cite{Purushotham}, \textbf{InfoVDANN} \cite{TuYouzhi}, and \textbf{DPTR} \cite{ChaiTNNLS}.
Table \ref{tab:TEP_setpoints} shows six different operation modes of the TEP dataset corresponding to different mass ratios and production rates. 
Each experiment follows a cross-domain validation strategy in which one operating mode is designated as the \textbf{target domain} (unlabeled), and the remaining five serve as the \textbf{source domain} (labeled) for training.
This procedure is repeated six times so that each mode serves once as the target, providing a comprehensive evaluation of unsupervised cross-domain soft sensing across different operating modes. The remaining hyperparameters follow the same configurations as described in the MFP experimental setup.

\subsection{TEP Results and Discussions}

As shown in Table \ref{tab:TEP_CaseStudies_results}, the proposed DVPF framework demonstrates consistent performance across all transfer tasks, maintaining high R2 and low RMSE and MAE scores. For instance, the M1, M2, M3, M4, M6 $\rightarrow$ M5 task achieves the best performance, highlighting the model's capability in performing soft sensing in unseen operating modes. Even in more challenging tasks, such as M1, M3, M4, M5, M6 $\rightarrow$ M2, where mode 2 exhibits greater statistical variability, DVPF still maintains overall regression metrics higher than all baselines. These results confirm that the DVPF excels in capturing nonlinear cross-mode dependencies and adapts to unseen process distributions without relying on target labels.
Fig. \ref{fig:tsne_tep} shows the t-SNE embeddings of the learned latent features for the first TEP transfer task. Consistent with the observations from the MFP case study, the DVPF demonstrates improved alignment between domains, with reduced overlap across operating regimes. In addition, the DVPF-learned latent features exhibit more compact intra-domain structure and smoother latent manifolds compared with the baseline methods, indicating improved representation consistency under heterogeneous operating conditions. Similarly, these results suggest that the DVPF effectively captures transferable process representations that remain stable under substantial domain shifts in complex industrial processes. Nevertheless, due to the higher process complexity and more heterogeneous operating conditions of the TEP, preservation of the ground-truth embedding structure is less pronounced than in the MFP.

\begin{table*}[t]
\centering
\caption{Computational efficiency comparison on the TEP case study. $B$ denote batch size, $T$ sequence length, $N$ number of particles used during training, 
$d_z$ latent dimension, $d_h$ RNN hidden dimension, $d_x$ measurement dimension, and $d_y$ label dimension.
$c_{\text{prior}}$, $c_{\text{enc}}$, $c_{\text{dec}}$, $c_{\phi}$, $c_{\text{adv}}$ denote the per-evaluation multiply-add time complexity of the prior, encoder, decoder, potential, and adversarial MLPs, respectively. $a_{\text{prior}}$, $a_{\text{enc}}$, $a_{\text{dec}}$, $a_{\phi}$, $a_{\text{adv}}$ denote the corresponding activation storage. 
A forward pass of $K$-layer MLP with widths $\{m_l\}_{l=0}^{K}$ incurs time complexity $c = \sum_{l=1}^{K} m_{l-1} m_l + m_l$ and space complexity $a = \sum_{l=1}^{K-1} m_l$. 
}
\label{tab:computational_efficiency}
\resizebox{\textwidth}{!}{%
\setlength\extrarowheight{0.1cm}
\begin{tabular}{@{}lcccccc@{}}
\toprule
Model & DARNN & VRADA & DPTR & DVPF (ours) \\ 
\midrule
Time Complexity & $\mathcal{O}\big( B T N (d_h(d_x \!+\! d_h) \!+\! c_{\text{dec}} \!+\! c_{\text{adv}}) \big)$ & $\mathcal{O}\big( B T N \big(d_h(d_x \!+\! d_z \!+\! d_h) \!+\! c_{\text{enc}} \!+\! c_{\text{prior}} \!+\! c_{\text{dec}} \!+\! d_x\big) \!+\! B N c_{\text{adv}} \big)$ & $\mathcal{O}\big( B T ( c_\text{enc} \!+\! c_\text{prior} \!+\! d_z ) \!+\! B T N (c_\text{dec} \!+\! c_\text{adv} \!+\! d_z \!+\! d_y) \big)$ & $\mathcal{O}\big(B T N (d_h(d_z \!+\! d_h) \!+\! c_{\text{prior}} \!+\! c_{\text{dec}} \!+\! c_{\phi} \!+\! d_x)\big)$ \\ 
\midrule
Space Complexity & $\mathcal{O}\big( B T N (d_h \!+\! c_{\text{dec}} \!+\! c_{\text{adv}}) \big)$ & $\mathcal{O}\big( B T N \big(d_h \!+\! d_z \!+\! a_{\text{enc}} \!+\! a_{\text{prior}} \!+\! a_{\text{dec}}\big) \!+\! B N a_\text{adv} \big)$ & $\mathcal{O}\big( B T ( a_\text{enc} \!+\! a_\text{prior} \!+\! d_z ) \!+\! B T N (a_\text{dec} \!+\! a_\text{adv}) \big)$ & $\mathcal{O}\big(B T N (d_h \!+\! d_z \!+\! a_{\text{prior}} \!+\! a_{\text{dec}} \!+\! a_{\phi})\big)$ \\ 
\midrule
Training Wall-Clock Time (s) & $ 260.055$ & $459.603$ & $451.099$ & $346.918$ \\ 
\midrule
Inference Wall-Clock Time (s) & $195.250$ & $336.625$ & $265.853$ & $248.968$ \\ 
\bottomrule
\end{tabular}%
}
\end{table*}

\subsection{Computational Efficiency Analysis}

We evaluate the computational cost of the proposed method by analyzing both the theoretical time and space complexity and the empirical training and inference wall-clock runtime averaged across six transfer tasks. As summarized in Table \ref{tab:computational_efficiency}, the proposed DVPF framework exhibits competitive computational efficiency, achieving the second-best empirical runtime among the (domain) adversarial-based methods. This observation is consistent with the theoretical analysis, which shows that the DVPF scales linearly with the batch size, sequence length, and number of particles, while excluding both the adversarial discriminator and the encoder networks required by several baseline methods.
Overall, these results demonstrate that the DVPF achieves a favorable trade-off between modeling expressiveness and computational efficiency. By avoiding adversarial training loops and encoder-based inference, and by employing a single-pass potential flow Bayesian update together with an efficient RNN parameterization and a tractable variational potential loss, the DVPF not only delivers strong predictive performance but also remains computationally feasible for industrial soft sensing applications.

\section{Conclusion}

In this paper, we propose a novel DVPF framework for unsupervised cross-domain soft sensor modeling. The framework consists of an RNN-parameterized sVBO that reframes cross-domain soft sensing as a sequential variational inference problem, followed by a potential flow that performs Bayesian inference to obtain a domain-adaptable posterior feature representation. 
The superior performance of the DVPF against existing deep feature-based domain adaptation and normalizing flows approaches provides strong evidence for the effectiveness of our framework in addressing cross-domain soft sensing modeling under incomplete sensor labels. 

\bibliographystyle{IEEEtran}
\bibliography{ref}

\end{document}